\documentclass[10pt,twocolumn,letterpaper]{article}

\usepackage{cvpr}
\usepackage{times}
\usepackage{epsfig}
\usepackage{graphicx}
\usepackage{amsmath}
\usepackage{amssymb}
\usepackage{authblk}

\usepackage{mmstyles}
\usepackage{siunitx}
\usepackage{booktabs}
\usepackage{subfigure}

\graphicspath{{figures/}}
\DeclareGraphicsExtensions{.pdf}

\usepackage[pagebackref=true,breaklinks=true,letterpaper=true,colorlinks,bookmarks=false]{hyperref}

\cvprfinalcopy 

\def\cvprPaperID{2957} 

\ifcvprfinal\fi
\begin{document}

\title{Online Deep Clustering for Unsupervised Representation Learning}

\author{Xiaohang Zhan\thanks{Equal Contribution.}~~\textsuperscript{1}, Jiahao Xie$^{*}$\textsuperscript{2}, Ziwei Liu\textsuperscript{1}, Yew Soon Ong\textsuperscript{2,3}, Chen Change Loy\textsuperscript{2} \\
\textsuperscript{1}CUHK - SenseTime Joint Lab, The Chinese University of Hong Kong\\
\textsuperscript{2}Nanyang Technological University \quad \textsuperscript{3}AI3, A*STAR, Singapore\\~ \\
\textsuperscript{1}\tt\small \{zx017, zwliu\}@ie.cuhk.edu.hk \\
\textsuperscript{2}\tt\small \{jiahao003, asysong, ccloy\}@ntu.edu.sg}

\maketitle
\thispagestyle{empty}


\begin{abstract}
Joint clustering and feature learning methods have shown remarkable performance in unsupervised representation learning. However, the training schedule alternating between feature clustering and network parameters update leads to unstable learning of visual representations. To overcome this challenge, we propose Online Deep Clustering (ODC) that performs clustering and network update simultaneously rather than alternatingly. Our key insight is that the cluster centroids should evolve steadily in keeping the classifier stably updated. Specifically, we design and maintain two dynamic memory modules, \ie, samples memory to store samples' labels and features, and centroids memory for centroids evolution. We break down the abrupt global clustering into steady memory update and batch-wise label re-assignment. The process is integrated into network update iterations. In this way, labels and the network evolve shoulder-to-shoulder rather than alternatingly. Extensive experiments demonstrate that ODC stabilizes the training process and boosts the performance effectively.
Code: \url{https://github.com/open-mmlab/OpenSelfSup}.
\end{abstract}


\section{Introduction}
%
%

\begin{figure}[t]
	\centering
	\includegraphics[width=\linewidth]{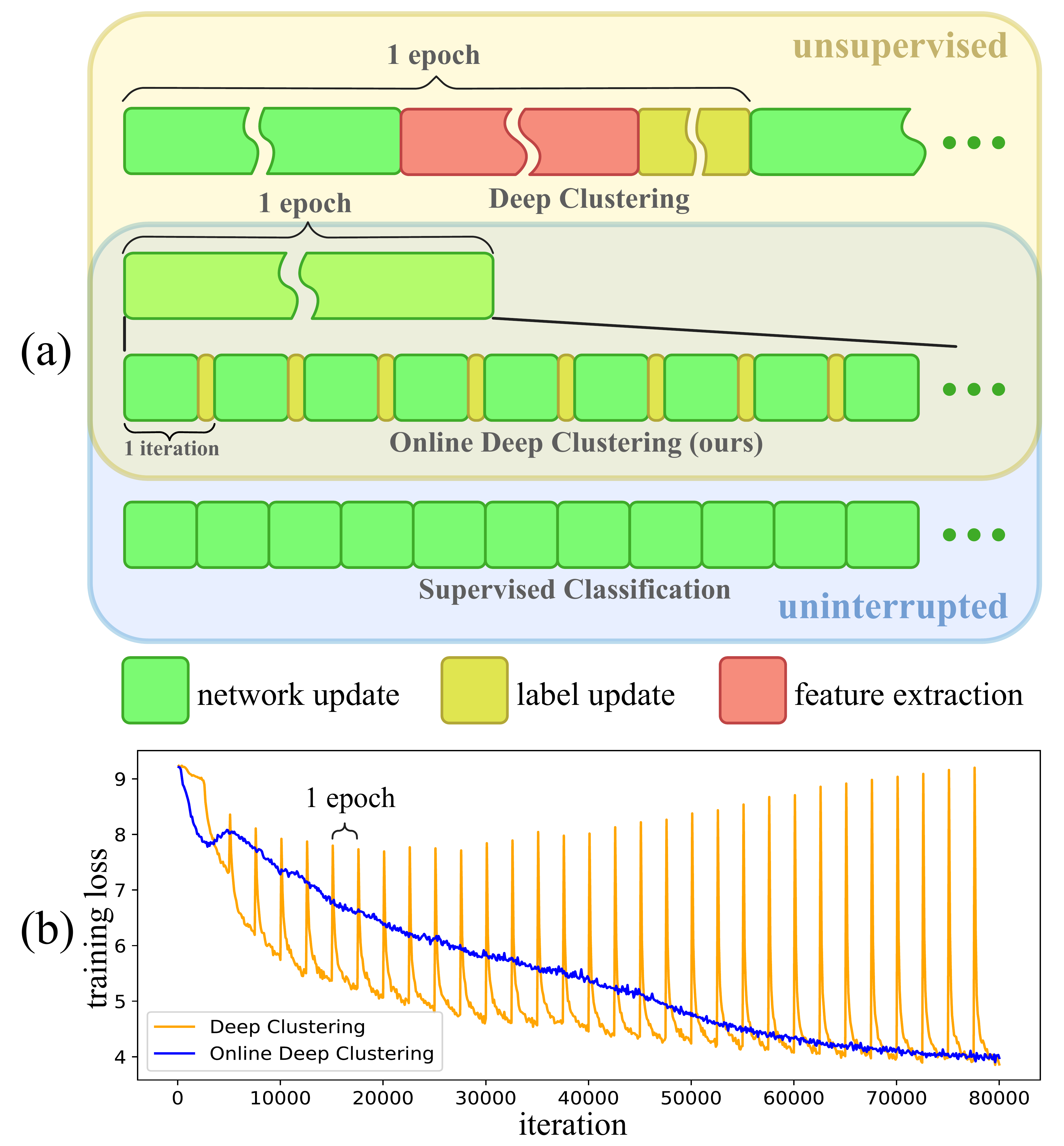}
	\caption{(a) Online Deep Clustering (ODC) seeks to reduce the discrepancy in training mechanism between Deep Clustering (DC) and supervised classification via integrating clustering process into network update iterations. ODC training is both unsupervised and uninterrupted. (b) Compared to DC, ODC updates labels continuously rather than in a pulsating manner, enabling the representations to evolve steadily. The loss curves (only initial 32 epochs for clarity) show the stability of ODC. After training, the loss is decreased to around 2.0 for ODC while 2.9 for DC.}
	\label{fig:teaser}
\end{figure}

Unsupervised representation learning~\cite{doersch2015unsupervised,pathak2016context,zhang2016colorful,noroozi2016unsupervised,donahue2017adversarial,larsson2017colorization,jenni2018self,gidaris2018unsupervised,zhan2019self} aims at learning transferable image or video representations without manual annotations.
Among them, clustering-based representation learning methods~\cite{huang2016unsupervised,xie2016unsupervised,yang2016joint,caron2018deep,caron2019unsupervised} emerge as a promising direction in this area.
Different from recovering-based approaches~\cite{pathak2016context,zhang2016colorful,noroozi2016unsupervised,gidaris2018unsupervised}, clustering-based methods require little domain knowledge~\cite{caron2018deep} while achieving encouraging performances.
Compared to contrastive representation learning~\cite{henaff2019data,he2019momentum,chen2020simple} that captures merely intra-image invariance, clustering-based methods are able to explore inter-image similarity.
Unlike conventional clustering that is typically performed on fixed features~\cite{zhan2018consensus,yang2019learning}, these works jointly optimize clustering and feature learning.
%
%
%
%

While evaluations of early works~\cite{xie2016unsupervised,yang2016joint} are mostly performed on small datasets, Deep Clustering~\cite{caron2018deep} (DC) proposed by Caron~\etal is the first attempt to scale up clustering-based representation learning.
DC alternates between deep feature clustering and CNN parameters update.
In particular, at the start of each epoch, it performs off-line clustering algorithms on the entire dataset to obtain pseudo-labels as the supervision for the next epoch.
Off-line clustering inevitably permutes the assigned labels in different epochs, \ie, even if some of the clusters do not change, their indices after clustering will be permuted randomly.
As a result, parameters in the classifier cannot be inherited from the last epoch and they have to be randomly initialized before each epoch.
The mechanism introduces training instability and exposes representations to a high risk of representation corruption. 
%
%
%
As shown in Figure~\ref{fig:teaser} (a), network update in DC is interrupted by feature extraction and clustering in each epoch.
This is in contrast to the conventional supervised classification that is performed in an uninterrupted manner using fixed labels, where an iteration consists of forward and backward propagations of the network.

In this work, we seek to devise a joint clustering and feature learning paradigm with high stability.
To reduce the discrepancy of training mechanism between DC and supervised learning, we decompose the clustering process into mini-batch-wise label update, and integrate this update process into iterations of network update.
Based on this intuition, we propose Online Deep Clustering (ODC) for joint clustering and feature learning.
Specifically, an ODC iteration consists of forward and backward propagations, label re-assignment, and centroids update.
For label update, ODC reuses the features in the forward propagation, thus avoiding additional feature extraction.
To facilitate online label re-assignment and centroids update, we design and maintain two dynamic memory modules, \ie, samples memory to store samples' labels and features, and centroids memory for centroids evolution.
In this way, ODC is trained in an uninterrupted manner similar to supervised classification, while no manual annotation is required.
During the training process, labels and network parameters evolve shoulder-to-shoulder, rather than alternatingly.
Since labels are updated in each iteration continuously and instantly, the classifier in the CNN also evolves more steadily, resulting in a much more steady loss curve as shown in Figure~\ref{fig:teaser} (b).

%
%
%
%
%
%


While ODC alone achieves compelling unsupervised representation learning performance on various benchmarks, it can be naturally used to fine-tune models that have been trained using other unsupervised learning approaches.
Extensive experiments show that the steadiness of ODC helps it to perform superiorly over DC as an unsupervised fine-tuning tool.
We conclude our contributions as follows:
{\bf 1)} we propose ODC that learns image representations in an unsupervised manner with high stability.
{\bf 2)} ODC also serves as a unified unsupervised fine-tuning scheme that further improves previous self-supervised representation learning approaches.
{\bf 3)} Promising performances are observed on different benchmarks, indicating the great potential of joint clustering and feature learning.

\section{Related Work}

\noindent\textbf{Unsupervised Representation Learning.}
Many unsupervised visual representation learning algorithms are based on generative models, which usually use a latent representation bottleneck to reconstruct input images.
Existing generation-based models include Auto-Encoders~\cite{vincent2008extracting,le2013building}, Restricted Boltzman Machines~\cite{hinton2006fast,lee2009convolutional,tang2012robust}, Variational Auto-Encoders~\cite{kingma2013auto} and Generative Adversarial Networks~\cite{goodfellow2014generative}, some of which have shown powerful ability in generating images or videos~\cite{denton2015deep,ledig2017photo,zhang2017stackgan,brock2018large,vondrick2016generating,tulyakov2018mocogan}.
By learning to generate examples, these models can learn meaningful latent representations that can be used for downstream tasks~\cite{donahue2017adversarial,dumoulin2016adversarially,donahue2019large}.

Another popular form of unsupervised representation learning is self-supervised learning, where a pretext task is designed to derive proxy labels from raw data.
Representations are learned by encouraging a CNN to predict the proxy labels from the data.
Various pretext tasks have been explored, \eg, predicting relative patch locations within an image~\cite{doersch2015unsupervised}, solving jigsaw puzzles~\cite{noroozi2016unsupervised}, colorizing grayscale images~\cite{zhang2016colorful,larsson2016learning}, inpainting of missing pixels~\cite{pathak2016context}, cross-channel prediction~\cite{zhang2017split}, counting visual primitives~\cite{noroozi2017representation}, predicting image rotations~\cite{gidaris2018unsupervised}, and multiview contrastive learning~\cite{tian2019contrastive}.
For videos, self-derived supervision signals come from temporal continuity~\cite{mobahi2009deep,jayaraman2016slow,wang2015unsupervised,liu2017video,lee2017unsupervised,misra2016shuffle,wei2018learning} or motion consistency~\cite{pathak2017learning,mahendran2018cross,walker2015dense,walker2016uncertain,zhan2019self}.

\begin{figure*}[t]
	\centering
	\includegraphics[width=.9\linewidth]{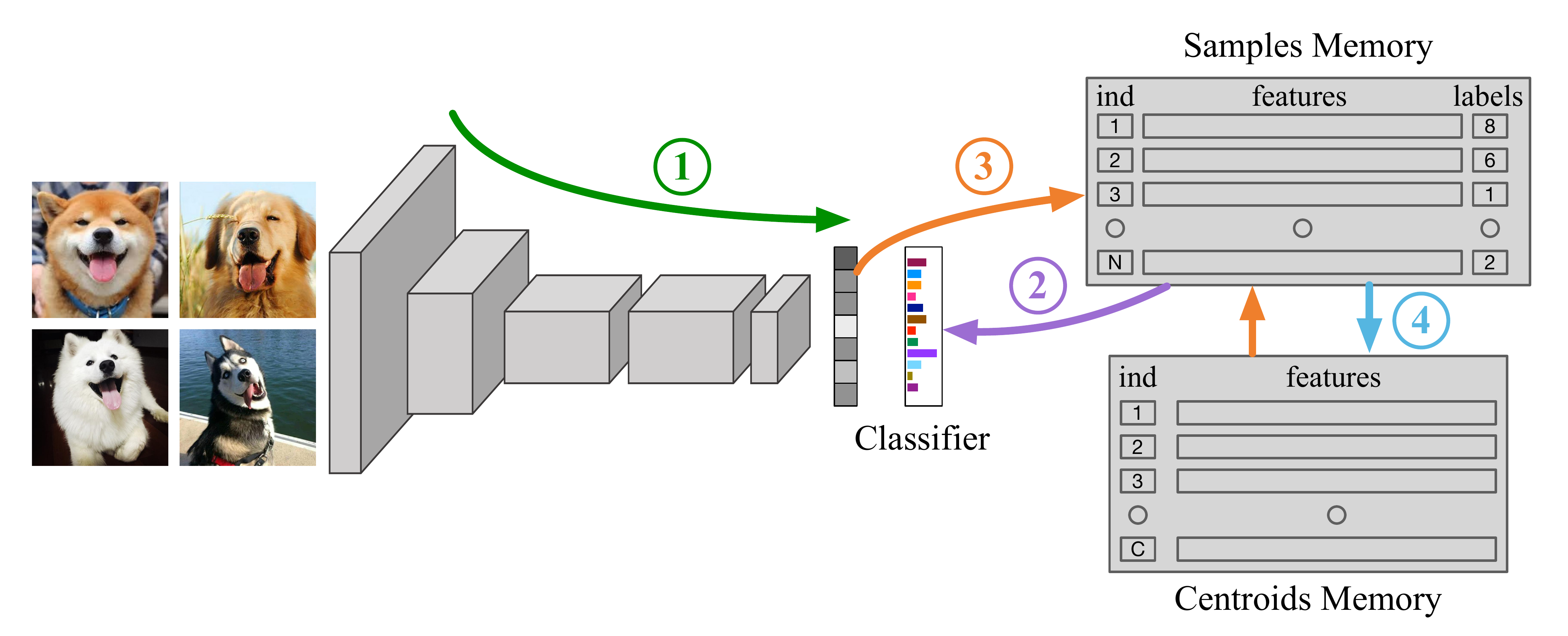}
	\caption{Each ODC iteration mainly contains four steps: 1. forward to obtain a compact feature vector; 2. read labels from the samples memory and perform back-propagation to update the CNN; 3. update samples memory by updating features and assigning new labels; 4. update centroids memory by recomputing the involved centroids.}
	\label{fig:framework}
	\vspace{0.25cm}
\end{figure*}

\noindent\textbf{Joint Clustering and Feature Learning.}
Clustering-based unsupervised representation learning is of particular interest recently.
Various methods are proposed to jointly optimize feature learning and image clustering. Notably, these methods have shown great potential in learning unsupervised features on small datasets~\cite{xie2016unsupervised,yang2016joint,dosovitskiy2014discriminative,liao2016learning}.
To scale up to large datasets like ImageNet~\cite{deng2009imagenet}, Caron~\etal~\cite{caron2018deep} propose DeepCluster to cluster features and update CNN with subsequent assigned pseudo-labels for each epoch.
In a subsequent study, Caron~\etal~\cite{caron2019unsupervised} propose DeeperCluster to leverage self-supervision and clustering, and validate the representation learning ability of their approaches on non-curated data.
Although deep clustering methods are capable of learning good representations from large-scale unlabeled data, the alternating update of feature clustering and CNN parameters update leads to instability in training.

\noindent\textbf{Improvements to Self-supervised Learning.}
Some works aim at improving previous self-supervised learning approaches from different perspectives.
For instance, Larsson~\etal~\cite{larsson2017colorization} give a first in-depth analysis on colorization as a pretext task and provide some insights on improving its effectiveness.
Mundhenk~\etal~\cite{mundhenk2018improvements} explore a set of methods to avoid some trivial shortcuts like chromatic aberration on context-based self-supervised learning.
Noroozi~\etal~\cite{noroozi2018boosting} improve the performance of self-supervised models using a clustering-based knowledge transfer method that allows a deeper network during pre-training.
Wang~\etal~\cite{wang2017transitive} and Doersch~\etal~\cite{doersch2017multi} exploit multiple cues contained in different pretext tasks to improve self-supervised models.
Recently, some works~\cite{kolesnikov2019revisiting,goyal2019scaling} have studied extensively the architectures and scaling ability on existing self-supervised approaches.
Complementary to these works, ODC serves as a flexible and unified unsupervised fine-tuning scheme to boost general self-supervised learning methods although it can be used alone to perform unsupervised representation learning from scratch.
%

\section{Methodology}

In the following sub-sections, we first discuss the differences between the proposed ODC to the conventional DC~\cite{caron2018deep} in Sec.~\ref{subsec:odc}.
We then recommend some useful strategies to maintain stable cluster size while using ODC in Sec.~\ref{subsec:cluster}. We finally explain how one can use ODC for unsupervised fine-tuning (Sec.~\ref{subsec:finetune}) and the implementation details of ODC (Sec.~\ref{subsec:details}).

\subsection{Online Deep Clustering}
\label{subsec:odc}

We first discuss the basic idea of DC~\cite{caron2018deep} and then detail the proposed ODC.
To learn representations, DC alternates between off-line feature clustering and network back-propagation with pseudo-labels.
The off-line clustering process requires deep feature extraction on the entire training set, followed by a global clustering algorithm, \eg, K-Means clustering.
The global clustering permutes the pseudo labels vastly, requiring the network to adapt to new labels rapidly in the subsequent epoch.

\noindent\textbf{Framework Overview.}
Different from DC, ODC does not require an extra feature extraction process. Besides, labels evolve alongside the network parameters update smoothly.
This is made possible by the newly introduced samples and centroids memories.
As shown in Fig.~\ref{fig:framework}, the samples memory stores features and pseudo-labels of the entire dataset; while the centroids memory stores the features of class centroids, \ie, the mean feature of all samples in a class.
A ``class'' here represents a temporary cluster that evolves continuously during training.
Labels and network parameters are updated simultaneously during uninterrupted iterations of ODC.
Specific techniques including \emph{loss re-weighting} and \emph{dealing with small clusters} are introduced to avoid ODC from getting stuck into trivial solutions.

\noindent\textbf{An ODC Iteration.} Assuming that we are given with a randomly initialized network $f_\theta\left(*\right)$ along with a linear classifier $g_w\left(*\right)$, the goal is to train the backbone parameters $\theta$ to produce highly discriminative representations.
To prepare for ODC, the samples and centroids memories are initialized via a global clustering process, \eg, K-Means.
Next, one can perform uninterrupted ODC iteratively.

An ODC iteration contains four steps.
First, given a batch of input images $\left\{x\right\}$, the network maps the images into compact feature vectors $F = f_\theta\left(x\right)$.
Second, we read pseudo-labels for this batch from the samples memory. With the pseudo-labels, we update the network with stochastic gradient descent to solve the following problem:
\begin{equation}
\min_{\theta,w} \frac{1}{B}\sum_{n=1}^{B}l\left(g_w\left(f_\theta\left(x_n\right)\right), y_n\right),
\end{equation}
where $y_n$ is the current pseudo label from the samples memory, $B$ denotes the size of each mini-batch.
Third, $f_\theta\left(x\right)$ after L2 normalization is reused to update the samples memory:
\begin{equation}
F_m\left(x\right) \leftarrow m \frac{f_\theta\left(x\right)}{\left \| f_\theta\left(x\right) \right \|_2} + \left(1-m\right) F_m\left(x\right),
\end{equation}
where $F_m\left(x\right)$ is the feature of $x$ in the samples memory, $m\in \left(0,1\right]$ is a momentum coefficient.
Simultaneously, each involved sample is assigned with a new label by finding the nearest centroid following:
\begin{equation}
\min_{y\in\left\{1,..,C\right\}}\left\|F_m\left(x\right) - C_y\right\|_2^2,
\end{equation}
where $C_y$ denotes the centroid feature of class $y$.
Finally, the involved centroids, including those in which new members join, and those from which old members leave, are recorded.
They are updated every $k$-th iterations by averaging the features of all samples belonging to their corresponding centroid.

\subsection{Handling Clustering Distribution in ODC}
\label{subsec:cluster}

\noindent\textbf{Loss Re-weighting.}
To avoid the training from collapsing into a few huge clusters, DC adopts uniform sampling before each epoch.
However, for ODC, the number of samples over the clusters changes in each iteration.
Using uniform sampling requires one to re-sample the entire dataset in each iteration, a process that is deemed redundant and costly.
We propose an alternative approach, \ie, re-weighting the loss according to the number of samples in each class.
To verify their equivalence, we implement a DC model with loss re-weighting and empirically find that the performance remains unchanged when the weight follows $w_c\propto \frac{1}{\sqrt{N_c}}$, where $N_c$ denotes the number of samples in class $c$.
Hence, we adopt the same loss re-weighting formulation for ODC.
With loss re-weighting, samples in smaller clusters contribute more towards backpropagation, thus pushing the decision boundary farther to accept more potential samples.

\noindent\textbf{Dealing with Small Clusters.}
Loss re-weighting helps to prevent the formation of huge clusters. Nevertheless, we still face the risk of having some small clusters collapsing into empty clusters.
To overcome this problem, we propose to process and eliminate extremely small clusters in advance before they collapse.
Denoting normal clusters as $\mathcal{C}_n$ whose sizes are larger than a threshold, and small clusters as $\mathcal{C}_s$ whose sizes are not, for $c\in \mathcal{C}_s$, we first assign samples in $c$ to the nearest centroids in $C_n$ to make $c$ empty.
Next, we split the largest cluster $c_{max} \in \mathcal{C}_n$ into two sub-clusters by K-Means and randomly choose one of the sub-clusters as the new $c$.
We repeat the process until all clusters belong to $\mathcal{C}_n$.
Though this process alters some clusters abruptly, it only affects a small portion of samples which are involved in this process.

\noindent\textbf{Dimensionality Reduction.}
Some of the backbone networks map an image to a high-dimensional vector, \eg, AlexNet produces 4,096-dimensional features and ResNet-50 yields 2,048-dimensional features, leading to high space and time complexities in subsequent clustering.
DC performed PCA on features across the entire dataset to reduce dimension.
However, for ODC, the features of different samples have varying timestamps, leading to incompatible statistics among samples.
Hence, PCA is not applicable anymore.
It is also costly to perform PCA in each iteration.
We therefore add a non-linear head layer of \{fc-bn-relu-dropout-fc-relu\} to reduce high dimensional features into 256 dimensions. It is jointly tuned during ODC iterations.
The head layer is removed for downstream tasks.

\subsection{ODC for Unsupervised Fine-tuning}
\label{subsec:finetune}
Compared with self-supervised learning approaches that tend to capture intra-image semantics, clustering-based methods focus more on inter-image information.
Hence, DC and ODC are naturally complementary to previous self-supervised learning approaches.
As DC and ODC are not restricted to a specifically designed objective, like rotation angle or color prediction, they readily serve as an unsupervised fine-tuning scheme to boost the performance of existing self-supervised approaches.
In this paper, we study the effectiveness of DC and ODC as a fine-tuning process with initialization from different self-supervised learning methods.

\subsection{Implementation Details}
\label{subsec:details}
\noindent\textbf{Data Pre-processing.}
We use ImageNet that contains 1.28M images without labels for training.
Images are first randomly cropped to have a resolution of 224x224 with augmentation including random flipping and rotation ($\pm\ang{2}$).
DC adopts a Sobel filter on the images to avoid exploiting color as the shortcut.
Such a pre-processing step requires the downstream tasks to include the Sobel layer as well, which potentially limit its application.
We find that strong color jittering shows the same effect as a Sobel filter in avoiding shortcuts, while it allows normal RGB images as inputs.
Specifically, we adopt PyTorch style color jitter transform with brightness factor $\left(0.6,1.4\right)$, contrast factor $\left(0.6,1.4\right)$, saturation factor $\left(0,2\right)$, and hue factor $\left(-0.5,0.5\right)$.
Besides, we randomly convert images to grayscale with a probability of $0.2$.
The random color jittering and grayscale applied on training samples randomize the similarity measured in color. This discourages the network from exploiting trivial information from color.

\noindent\textbf{Training of ODC.}
We use ResNet-50 as our backbone.
Considering that most early works use AlexNet, we also perform experiments on AlexNet for comparison.
Following~\cite{caron2018deep}, we use AlexNet architecture without Local Response Normalization and add batch normalization layers.
The ODC models for AlexNet and ResNet-50 are trained from scratch.
The batch size is 512 allocated to 8 GPUs.
The learning rate is constantly $0.04$ for AlexNet and $0.06$ for ResNet-50 for 400 epochs, and decayed by $0.1$ for further 40 epochs.
Following DC, the number of clusters is set as 10,000, which is 10 times larger than the annotated number of classes of ImageNet.
The momentum coefficient $m$ is set as $0.5$.
The threshold to identify small clusters is set as 20. Varying this threshold does not affect the results significantly, provided that it does not exceed the average number of samples in a cluster.
The centroids memory is updated in every 10 iterations. The centroids update frequency constitutes a trade-off between learning efficacy and efficiency.
In our experiments, we observe that as long as the frequency is restricted to a reasonable range, the performance of ODC is not sensitive to it.
%

\section{Experiments}

\subsection{Evaluation on Unsupervised Representation}

\begin{table*}[t]
	\centering
	\small
	\caption{AlexNet linear classification on ImageNet and Places. We report top-1 center-crop accuracy. Numbers for other methods are obtained either from~\cite{zhang2017split} or from their original papers. The highest performance in each layer is in bold, and the second highest performance in each layer is underlined. SplitBrain and CMC have half the number of parameters.}
	\begin{tabular}{@{}lcccccccccc@{}}
		\toprule
		Method     & \multicolumn{5}{c}{ImageNet}     & \multicolumn{5}{c}{Places}      \\ \cmidrule(lr){2-6}\cmidrule(l){7-11}
		(AlexNet)    & conv1 & conv2 & conv3 & conv4 & conv5 & conv1 & conv2 & conv3 & conv4 & conv5 \\ \midrule
		Places labels~\cite{zhang2017split}  & -   & -   & -   & -   & -   & 22.1 & 35.1 & 40.2 & 43.3 & 44.6 \\
		ImageNet labels~\cite{zhang2017split} & 19.3 & 36.3 & 44.2 & 48.3 & 50.5 & 22.7 & 34.8 & 38.4 & 39.4 & 38.7 \\
		Random~\cite{zhang2017split}     & 11.6 & 17.1 & 16.9 & 16.3 & 14.1 & 15.7 & 20.3 & 19.8 & 19.1 & 17.5 \\ \midrule
		Context~\cite{doersch2015unsupervised}  & 16.2 & 23.3 & 30.2 & 31.7 & 29.6 & 19.7 & 26.7 & 31.9 & 32.7 & 30.9 \\
		ContextEncoder~\cite{pathak2016context} & 14.1 & 20.7 & 21.0 & 19.8 & 15.5 & 18.2 & 23.2 & 23.4 & 21.9 & 18.4 \\
		Jigsaw~\cite{noroozi2016unsupervised}     & 19.2 & 30.1 & 34.7 & 33.9 & 28.3 & 23.0 & 32.1 & 35.5 & 34.8 & 31.3 \\
		Colorization~\cite{zhang2016colorful}  & 13.1 & 24.8 & 31.0 & 32.6 & 31.8 & 22.0 & 28.7 & 31.8 & 31.3 & 29.7 \\
		SplitBrain~\cite{zhang2017split}   & 17.7 & 29.3 & 35.4 & 35.2 & 32.8 & 21.3 & 30.7 & 34.0 & 34.1 & 32.5 \\
		Counting~\cite{noroozi2017representation}    & 18.0 & 30.6 & 34.3 & 32.5 & 25.7 & \underline{23.3} & \textbf{33.9} & 36.3 & 34.7 & 29.6 \\
		NPID~\cite{wu2018unsupervised} & 16.8 & 26.5 & 31.8 & 34.1 & 35.6 & 18.8 & 24.3 & 31.9 & 34.5 & 33.6 \\ 
		Rotation~\cite{gidaris2018unsupervised}    & 18.8 & 31.7 & 38.7 & 38.2 & 36.5 & 21.5 & 31.0 & 35.1 & 34.6 & 33.7 \\
		DeepCluster~\cite{caron2018deep}   & 12.9 & 29.2 & 38.2 & 39.8 & 36.1 & 18.6 & 30.8 & 37.0 & 37.5 & 33.1 \\
		AET~\cite{zhang2019aet} & 19.2 & 32.8 & \underline{40.6} & 39.7 & 37.7 & 22.1 & 32.9 & \underline{37.1} & 36.2 & 34.7 \\
		Rot-Decouple~\cite{feng2019self} & \underline{19.3} & \underline{33.3} & \textbf{40.8} & \textbf{41.8} & \textbf{44.3} & 22.9 & 32.4 & 36.6 & 37.3 & \textbf{38.6} \\
		LA~\cite{zhuang2019local} & 14.9 & 30.1 & 35.7 & 39.4 & 40.2 & 17.1 & 32.2 & 36.5 & \underline{38.3} & \underline{37.8} \\
		CMC~\cite{tian2019contrastive} & 18.3 & \textbf{33.7} & 38.3 & 40.5 & \underline{42.8} & -   & -   & -   & -   & -    \\\midrule
		ODC (Ours) & \textbf{19.6} & 32.8 & 40.4 & \underline{41.4} & 37.3 & \textbf{24.0} & \underline{33.2} & \textbf{38.3} & \textbf{38.4} &35.5 \\ \bottomrule
	\end{tabular}
	\label{tab:alexnet}
	\vspace{0.5cm}
\end{table*}

\begin{table*}[t]
	\centering
	\small
	\caption{ResNet-50 linear classification on ImageNet and Places. We report top-1 center-crop accuracy. Numbers for methods with $^\ast$ and $^\dag$ are produced by third-party studies as cited, and by us, respectively. Numbers for other methods are taken from their original papers. The highest performance in each layer is in bold, and the second highest performance in each layer is underlined. CMC has half the number of parameters.}
	\begin{tabular}{@{}lcccccccccc@{}}
		\toprule
		Method     & \multicolumn{5}{c}{ImageNet}     & \multicolumn{5}{c}{Places}      \\ \cmidrule(lr){2-6}\cmidrule(l){7-11}
		(ResNet-50)   & conv1 & conv2 & conv3 & conv4 & conv5 & conv1 & conv2 & conv3 & conv4 & conv5 \\ \midrule
		Places labels~\cite{goyal2019scaling}$^\ast$  & -   & -   & -   & -   & -   & 16.7 & 32.3 & 43.2 & 54.7 & 62.3 \\
		ImageNet labels~\cite{goyal2019scaling}$^\ast$ & 11.6 & 33.3 & 48.7 & 67.9 & 75.5 & 14.8 & 32.6 & 42.1 & 50.8 & 52.5 \\
		Random~\cite{goyal2019scaling}$^\ast$     & 9.6  & 13.7 & 12.0 & 8.0  & 5.6  & 12.9 & 16.6 & 15.5 & 11.6 & 9.0  \\ \midrule
		Jigsaw~\cite{goyal2019scaling}$^\ast$     & 12.4 & 28.0 & \underline{39.9} & 45.7 & 34.2 & 15.1 & 28.8 & 36.8 & 41.2 & 34.4 \\
		Colorization~\cite{goyal2019scaling}$^\ast$  & 10.2 & 24.1 & 31.4 & 39.6 & 35.2 & 14.7 & 27.4 & 32.7 & 37.5 & 34.8 \\
		NPID~\cite{wu2018unsupervised}      & \textbf{15.3} & 18.8 & 24.9 & 40.6 & 54.0 & 18.1 & 22.3 & 29.7 & 42.1 & 45.5 \\
		Rotation~\cite{kolesnikov2019revisiting}$^{\ast}$ & \multicolumn{5}{c}{41.7 (best layer)}  & \multicolumn{5}{c}{38.1 (best layer)} \\
		BigBiGAN~\cite{donahue2019large} & \multicolumn{5}{c}{55.4 (best layer)} & \multicolumn{5}{c}{-} \\
		DeepCluster~\cite{caron2018deep}$^\dag$ & 14.4	& \underline{29.6}	& \underline{39.9}	& \underline{52.2}	& 50.3	& \underline{19.3}	& \underline{31.9}	& 39.0	& 46.1	& 43.6 \\
		LA~\cite{zhuang2019local} & 9.3 & 23.2 & 38.0 & 48.6 & \textbf{58.8} & 18.3 & 31.5 & \underline{39.2} & \underline{46.3} & \underline{49.1} \\
		CMC~\cite{tian2019contrastive} & \multicolumn{5}{c}{\underline{58.4} (best layer)} & \multicolumn{5}{c}{-} \\ \midrule
		ODC (Ours) & \underline{14.8}&	\textbf{31.6}&	\textbf{42.5}&	\textbf{55.7}&	57.6 & \textbf{21.4}&	\textbf{35.0} &	\textbf{41.3}&	\textbf{47.4}&	\textbf{49.3} \\ \bottomrule
	\end{tabular}
	\label{tab:resnet}
\end{table*}

After pre-training the ODC model, we evaluate the quality of unsupervised features on standard downstream tasks including ImageNet classification, Places205~\cite{zhou2014learning} classification, VOC2007~\cite{everingham2010pascal} SVM classification, and VOC2007 Low-shot classification.
We provide the details of each benchmark and show our competing results as follows.

\noindent\textbf{Re-implementation of Deep Clustering.}
Since the original paper of DC does not include ResNet-50, we implement a DC model with ResNet-50.
The DC model adopts the same data augmentations as ODC, except that DC applies a Sobel filter on images.
For fair comparisons, the training hyper-parameters of DC are the same as ODC except that we empirically find $lr=0.1$ is more suitable for DC.

\noindent\textbf{ImageNet Classification.}
Following the setup in Zhang~\etal~\cite{zhang2017split}, we keep the backbone including all convolution and batch normalization layers frozen, and train a 1000-way linear classifier on features from different depths of convolutional layers.
The features are mapped to around 9000 dimensions via average pooling.
We train all models for 100 epochs in total, using SGD with a momentum of 0.9 and batch size of 256.
The learning rate is initialized as 0.01, decayed by a factor of 10 after every 30 epochs.
Other hyper-parameters are set following Goyal~\etal~\cite{goyal2019scaling}.
We report top-1 center-crop accuracy on the official validation split of ImageNet. 

For AlexNet, as shown in Table~\ref{tab:alexnet}, ODC has a consistent improvement over DC in all conv layers, with the largest improvement (6.7\%) observed in conv1 layer.
The performance in conv1 layer surpasses the ImageNet pre-trained model.
%
%
With regard to the best-performing layer, ODC achieves 41.4\% on conv4 layer, outperforming the latest LA~\cite{zhuang2019local}, ranking only second to Rot-Decoupling~\cite{feng2019self}.
Though ODC does not outperform Rot-Decoupling in its best performing layer, it provides a complementary perspective to rotation based methods.

ODC also scales well with deeper architectures.
For ResNet-50, as shown in Table~\ref{tab:resnet},
ODC achieves 57.6\% center-crop accuracy in the conv5 layer, which is 5.4\% higher than the best performing layer of the re-implemented DC.
Compared with the concurrent state-of-the-art method LA~\cite{zhuang2019local}, our method produces competing results.
Though the result of conv5 is slightly lower than LA, ODC outperforms LA from conv1 to conv4 layers by large margins.
We observe a consistent performance increase from shallower layers to deeper layers, indicating that ODC makes full use of all residual layers.

\noindent\textbf{Places205 Classification.}
%
%
Following Zhang~\etal~\cite{zhang2017split}, to test the generalization ability on other domains, we also transfer the learned models to Places205 dataset that contains 2.45M images of 205 scene categories.
Similar to the experiments on ImageNet, we train a 205-way linear classifier on top of each frozen convolutional layer on the train split of Places205, and report top-1 center-crop accuracy on the standard validation split.
The evaluation setting and hyper-parameters are the same as those in the ImageNet classification task.

The results in Table~\ref{tab:alexnet} show that ODC with AlexNet as the backbone outperforms DC in all layers as well.
%
ODC surpasses all previous works on conv1, conv3 and conv4 layers.
Similar to the observation in the ImageNet classification task, ODC scales well on deeper architectures when it is transferred to Places205 with ResNet-50.
As shown in Table~\ref{tab:resnet}, in all layers, ODC surpasses all previous works, with the largest margin (3.1\%) to the runner-up observed in conv2 layer.
%
%
With regard to the best performing layer, ODC reaches 49.3\% center-crop accuracy in the conv5 layer, surpassing the re-implemented DC by 3.2\% in the respective best layer.
We observe the superiority of ODC in conv1 and conv2 layers over the supervised model using either Places labels or ImageNet labels.
%
%
The transfer performance of our method in the Places205 classification task indicates that representations learned by ODC can generalize well to different domains from ImageNet.

\noindent\textbf{VOC2007 SVM Classification.}
To further evaluate the generalization of learned features, we perform experiments on the VOC2007 transfer learning task that resembles real applications with smaller datasets.
Following~\cite{goyal2019scaling}, we train linear SVMs on features extracted from the frozen backbone on the ``trainval'' split of VOC2007 and evaluate on the test split.
We follow the same test setting and hyper-parameters used in~\cite{goyal2019scaling}, and report the best performing layers of different methods for ResNet-50.
The results in Table~\ref{tab:voc} show that ODC surpasses previous approaches by a significant margin on the VOC2007 SVM classification task.
With ODC, we achieve 78.2\% mAP performance, which is 9.1\% higher than DC.
However, We also note that there is still a significant 9.8\% performance gap between our ODC and the supervised model pre-trained with ImageNet labels, leaving room for further exploration.

\begin{figure}[t]
	\centering
	\includegraphics[width=\linewidth]{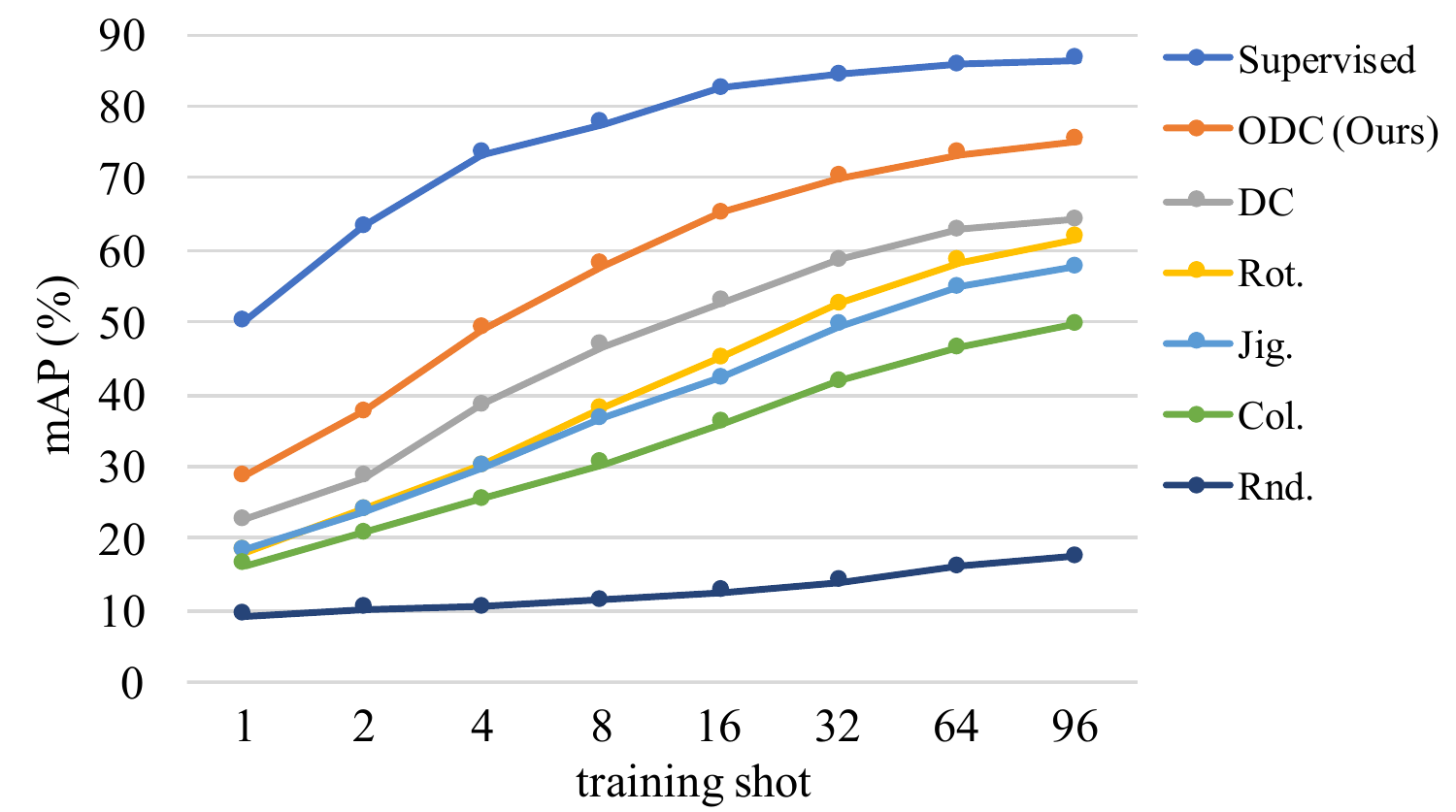}
	\caption{Low-shot Image Classification on VOC07 with linear SVMs trained and tested on the features from the best layer respectively for each method. We show the average performance for each shot across five runs.}
	\label{fig:lowshot}
\end{figure}

\noindent\textbf{Low-shot VOC2007 SVM Classification.}
Following~\cite{goyal2019scaling}, we also transfer our learned representations to a low-shot setting of VOC2007 SVM classification to test the quality of features when there are few training examples per category.
We vary the number of positive samples in each class and train linear SVMs on the frozen ResNet-50 backbone using the same setting from VOC2007 SVM classification.
We use the standard ``trainval'' split of VOC2007 in training and the test split in testing.
We report the mean average precision (mAP) across five independent samples for various low-shot values in Figure~\ref{fig:lowshot}.
The final mAP results shown in Table~\ref{tab:voc} are observed as the averages of all low-shot values and all independent runs.
The per-shot results are shown in Figure~\ref{fig:lowshot}. ODC has a consistent improvement over DC for each shot, with the performance gap further increasing when more positive examples per class are allowed.
We also observe that the performance gap between ODC and the supervised model pre-trained with ImageNet labels is gradually narrowed down with the increase of training shot values.
Table~\ref{tab:voc} shows that ODC achieves 57.1\% mAP performance in low-shot SVM classification on VOC2007, 10.2\% higher than our counterpart DC.
The low-shot results of ODC in this benchmark suggest that the learned features through ODC generalize well to low-shot classification.

\begin{table}[t]
\centering
\small
\caption{ResNet-50 SVM classification and low-shot SVM classification mAP on VOC07. Numbers for methods with$^\dag$ are produced by us. Numbers for other methods are taken from~\cite{goyal2019scaling}.}
\begin{tabular}{@{}lccc@{}}
\toprule
Method     & best & VOC07 SVM & VOC07 SVM \\
(ResNet-50)   & layer & (\% mAP) & Low-shot (\% mAP) \\ \midrule
ImageNet labels & 5 &88.0   & 75.4 \\
Random     & 1 &9.6    & 12.7 \\ \midrule
Jigsaw~\cite{noroozi2016unsupervised} & 4 & 64.5   & 39.2 \\
Colorization~\cite{zhang2016colorful} & 4 & 55.6   & 33.3\\
Rotation~\cite{gidaris2018unsupervised}$^\dag$ & 4 & 67.4  & 41.0 \\
DeepCluster~\cite{caron2018deep}$^\dag$ & 5 & 69.1  & 46.9 \\ \midrule
ODC (Ours) & 5 & \textbf{78.2} & \textbf{57.1} \\ \bottomrule
\end{tabular}
\label{tab:voc}
\end{table}

\subsection{Further Analysis}

In this section, we further analyze our ODC model from different perspectives.

\noindent\textbf{ODC as a Fine-tuning Scheme.}
The high efficiency enables ODC to easily serve as a rapid unsupervised fine-tuning scheme.
To assess the fine-tuning ability of ODC, we also use our reimplemented DC to fine-tune other self-supervised models.
The improvements over different self-supervised approaches are shown in Table~\ref{tab:finetuning}.
Compared with DC, we observe that ODC boosts the performance of each self-supervised approach by a significant margin.
With ODC fine-tuning, we achieve 16.7\% improvements for Col., 9.9\% for Jig., 7.1\% for Rot., and 7.9\% for DC, respectively, on the VOC2007 SVM classification benchmark.
By contrast, DC also yields fine-tuning improvements but lags far behind ODC.
%

\begin{table}[t]
\centering
\small
\caption{Improvements over previous self-supervised approaches. Each model is fine-tuned for 120 epochs. We report VOC07 SVM classification mAP for ResNet-50. Pre-trained models marked$^\ast$ are provided by~\cite{goyal2019scaling}, hence the original results are also taken from~\cite{goyal2019scaling}. For methods marked$^\dag$, we reimplement them to obtain the results.}
\resizebox{.48\textwidth}{!}{
\begin{tabular}{lccccc}
\toprule
     & Col.~\cite{zhang2016colorful}$^\ast$ & Jig.~\cite{noroozi2016unsupervised}$^\ast$ & Rot.~\cite{gidaris2018unsupervised}$^\dag$ & DC~\cite{caron2018deep}$^\dag$ \\ \midrule
Original & 55.6  & 64.5   & 67.4  & 69.1 \\
DC~\cite{caron2018deep}$^\dag$    & 61.2  & 68.5  & 68.6  & 70.0  \\
ODC   & \textbf{72.3}  & \textbf{74.4}  & \textbf{74.5}  & \textbf{77.0} & \\ \bottomrule
\end{tabular}
}
\label{tab:finetuning}
\end{table}

\begin{figure}[t]
	\centering
	\begin{tabular}{cc}
		\begin{minipage}[t]{0.5\linewidth}
			\includegraphics[width=\linewidth]{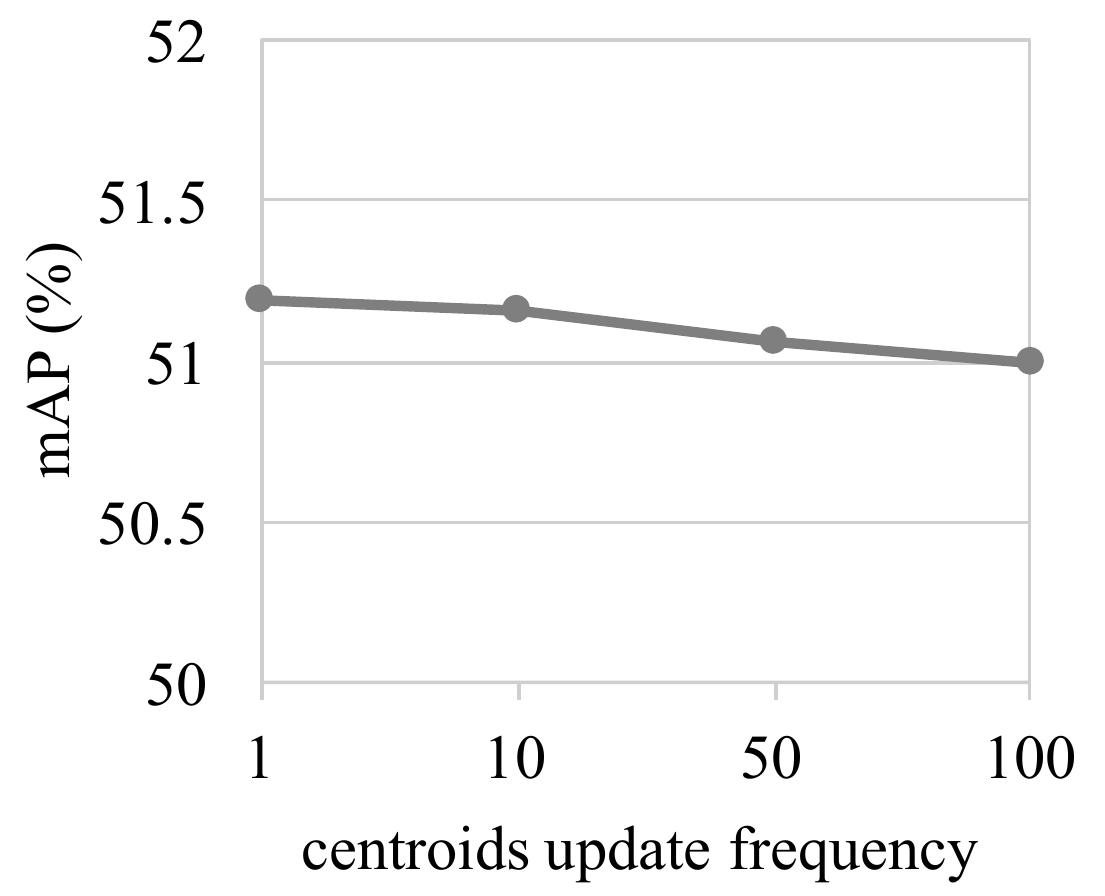}
		\end{minipage}
		\begin{minipage}[t]{0.5\linewidth}
			\includegraphics[width=\linewidth]{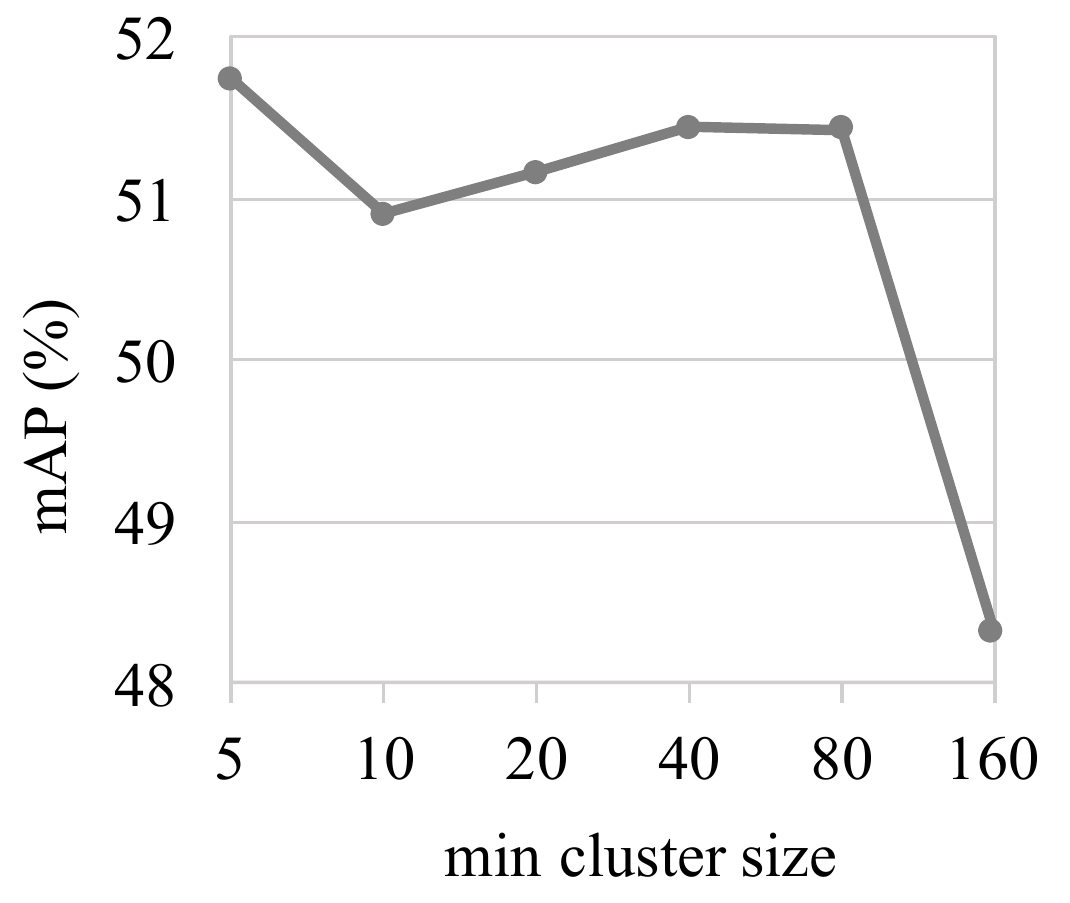}
		\end{minipage}
	\end{tabular}
	\caption{Influence of centroids update frequency (\textit{left}) and minimal small cluster size (\textit{right}) on the quality of features learned by ODC. We study these hyper-parameters on uniformly sampled 90K ImageNet within 300 random classes. We report mAP on VOC07 SVM classification task with ResNet-50.}
	\label{fig:hyperparams}
\end{figure}

\noindent\textbf{Influence of the Hyper-parameters.}
The hyper-parameters of ODC include the frequency of updating the centroids memory, and the minimal size of clusters.
To study the influence of the aforementioned two hyper-parameters, we train models with 90K images that are uniformly sampled from the original 1.28M ImageNet dataset, and evaluate the performance on VOC2007 SVM classification benchmark.
Figure~\ref{fig:hyperparams} shows the influence of the update frequency of centroids memory.
We observe no significant decrease in the performance of ODC when the update frequency becomes lower, indicating that our method is insensitive to this hyper-parameter provided that it is within a reasonable range.
The influence of the minimal size of small clusters is shown in Figure~\ref{fig:hyperparams}. The results show that a large threshold (~\ie 160) on clusters size would lead to a performance drop.
The result is not surprising. A cluster whose size is smaller than the minimal size is identified as a ``small cluster''. An overly frequent processing of such small clusters (see Sec.~\ref{subsec:cluster}) introduces instability in feature learning. The large threshold would also group images that should not have belonged to the same class.
It is noteworthy that ODC does not experience a significant change in performance within a reasonable range of minimal cluster sizes. 

\begin{figure}[t]
	\centering
	\includegraphics[width=.85\linewidth]{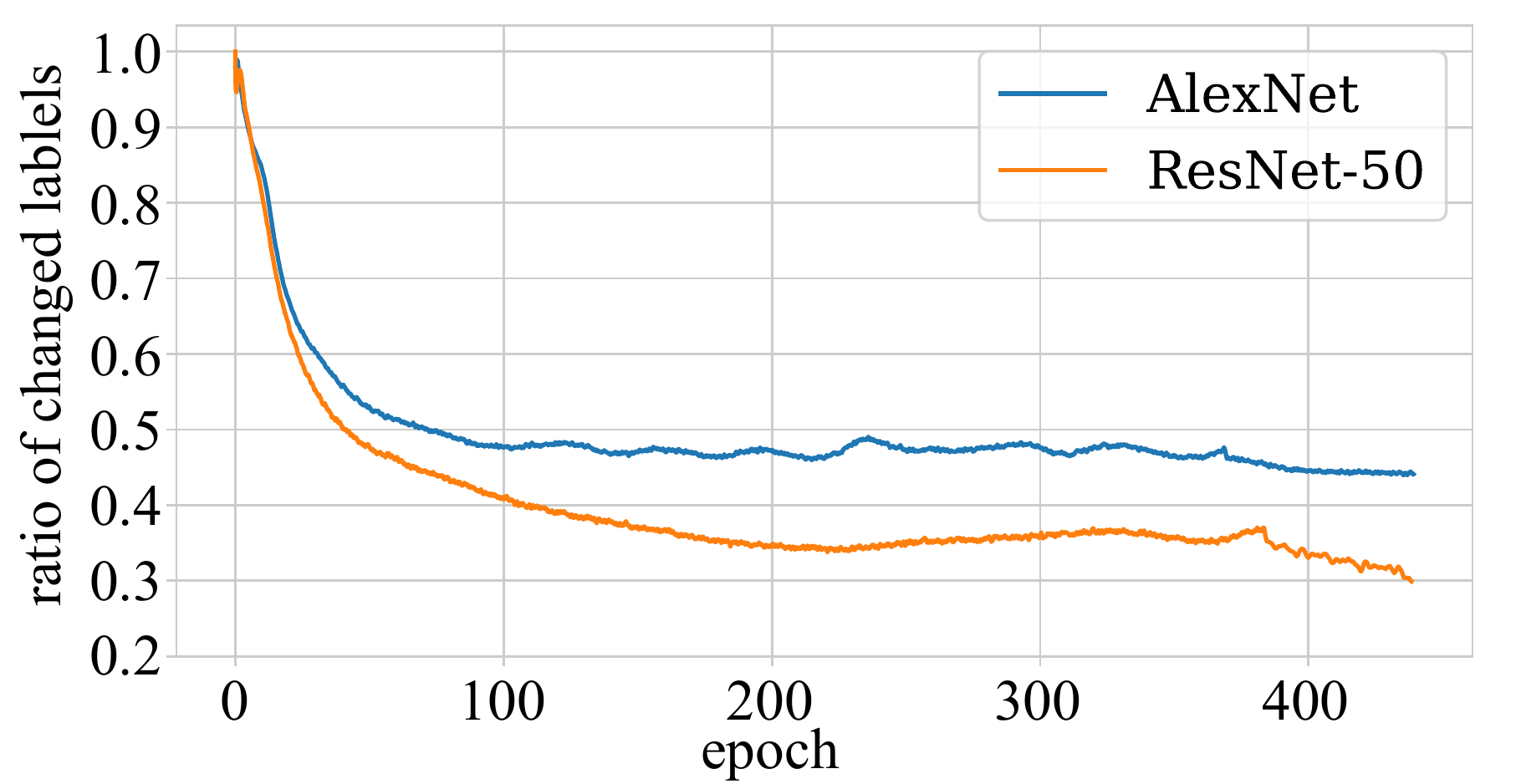}
	\caption{The ratio of changed labels in each batch gradually declines, indicating ODC tends to be stable during training.}
	\label{fig:changeratio}
\end{figure}

\noindent\textbf{Stability and Convergence.}
Figure~\ref{fig:teaser} already demonstrates the superior stability of ODC over DC from the aspect of the loss curve.
In Figure~\ref{fig:changeratio}, we show the training stability and convergence of ODC throughout the full training iterations.
To measure the stability of our models, we record the ratio of samples whose labels are changed in a batch. Intuitively, fewer label switchings suggest a higher stability. 
We report the ratio when different backbones are trained from scratch with ODC.
The curves begin with the highest label-switching ratio, \ie, nearly 100\% of samples in a batch experience a switch in their labels.
Gradually, the label-switching ratio declines and converges to a relatively low value.
Though there is always a small portion of samples altering their labels at last, ODC reaches a stable state.

\begin{figure}[t]
	\centering
	\includegraphics[width=\linewidth]{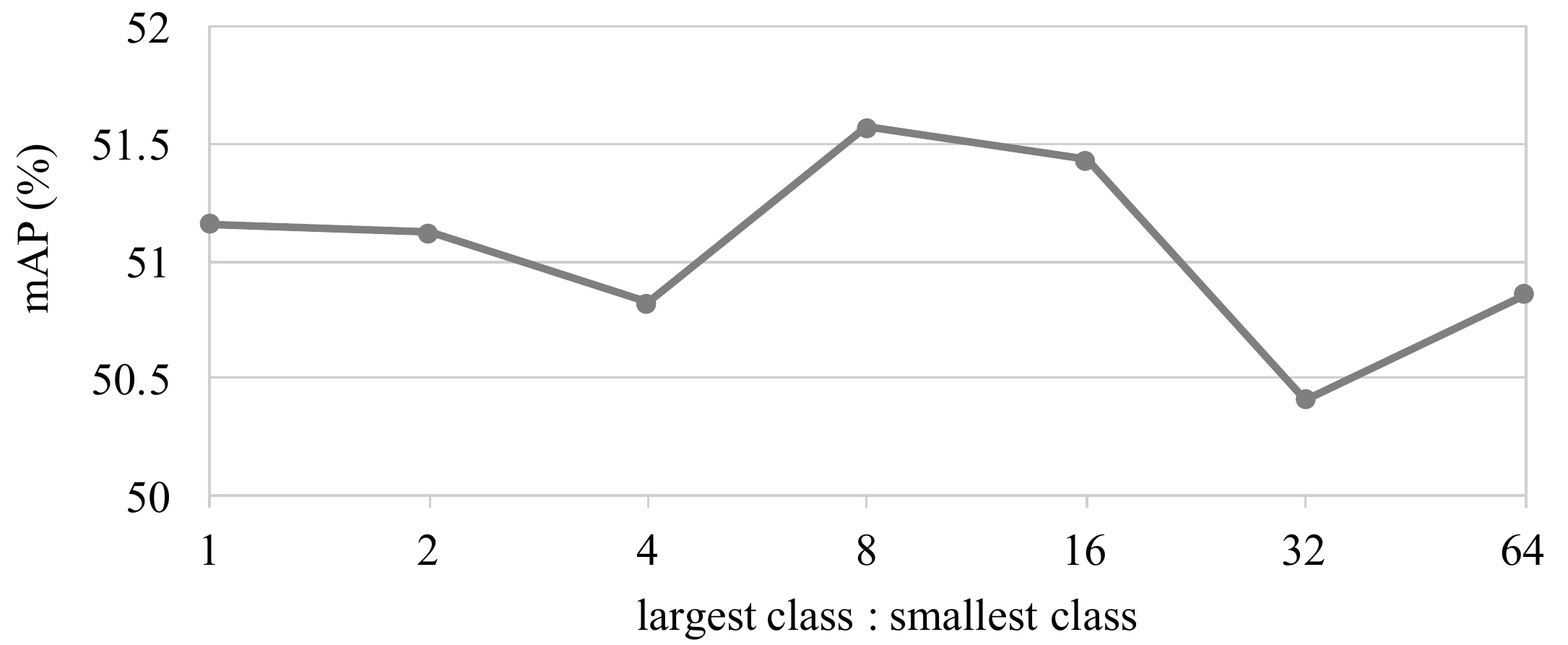}
	\caption{The efficacy of ODC trained on downsampled 300-class 100K long-tail ImageNet, with the ratio of the size of largest class to smallest class ranging from 1 (\textit{non-long-tail}) to 64 (\textit{highly long-tail}). We report mAP on VOC07 SVM task with ResNet-50.}
	\label{fig:longtail}
\end{figure}

\noindent\textbf{Training on Long-Tailed Data.}
In all previous experimens, we train our models on the class-balanced ImageNet dataset.
To evaluate the learning efficacy of ODC on long-tailed data, we perform experiments on downsampled long-tail ImageNet following~\cite{liu2019large}.
Specifically, we randomly downsample 300 classes with 100K images from the original ImageNet dataset to make different levels of long-tail ImageNet datasets, where the ratio of the largest class to the smallest class ranges from 1 (the non-long-tail level) to 64 (the highest long-tail level).
%
%
Figure~\ref{fig:longtail} shows the performance of ODC trained on different levels of long-tail ImageNet.
We observe no significant performance drop even in the conditions with large long-tail degrees, suggesting the robustness of our method on long-tailed data.
%

\begin{figure}[t]
	\centering
	\includegraphics[width=\linewidth]{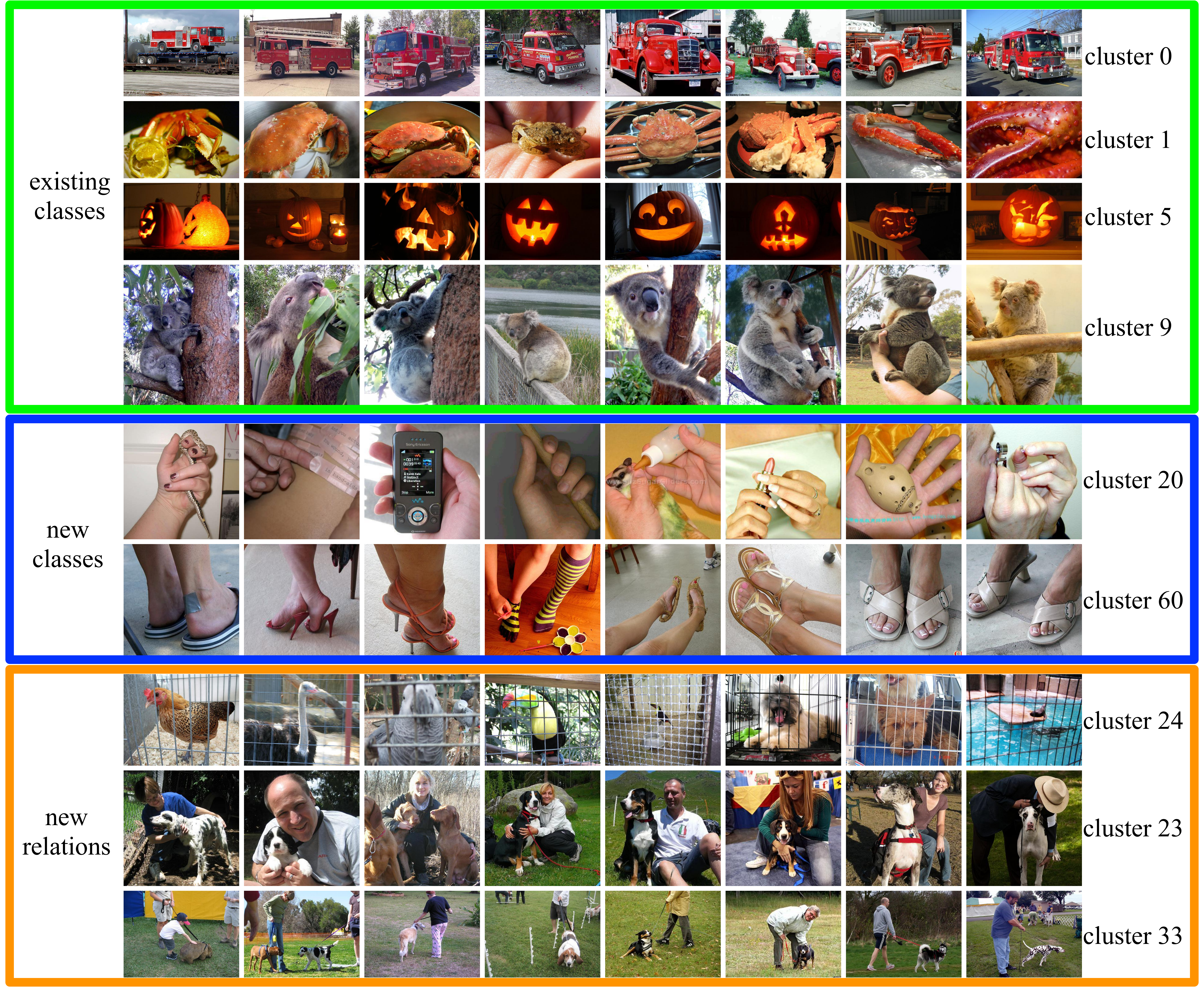}
	\caption{This figure shows part of selected clusters. Each row represents a cluster. Apart from the clusters that represents existing classes in ImageNet annotations, shown in the green box, we also find some new classes discovered by ODC. For example, the two rows in the blue box group ``hand'' and ``feet'' respectively, while ``hand'' or ``feet'' is not a category in ImageNet annotations. ODC also surprisingly groups images with similar relations between objects. As shown in the orange box, the clusters represent ``animal in cage'', ``person holds dog'' and ``person leads dog with a rope'' respectively.}
	\label{fig:cluster_vlz}
\end{figure}

\noindent\textbf{Visualization of Clusters.}
We visualize some selected clusters as shown in Figure~\ref{fig:cluster_vlz}.
Since the number of clusters is much larger than that of the original annotations, there will certainly be some clusters that represent new semantics beyond the annotated classes.
We find new classes, \eg, ``hand'' and ``feet'', and new relations, \eg, ``animal in cage'', ``person holds dog'' and ``person leads dog with a rope'', that are discovered by ODC.
The phenomenon reveals the potential of unsupervised learning to capture new semantics beyond manual annotations.

\section{Conclusion}
We have proposed an effective joint clustering and feature learning paradigm for unsupervised representation learning.
The proposed approach, Online Deep Clustering (ODC), attains effective and stable unsupervised training of deep neural networks, via decomposing feature clustering and integrating the process into iterations of network update.
ODC performs compellingly as an unsupervised representation learning scheme alone. It can also be used to fine-tune and substantially improve previous self-supervised learning methods.
%
%

\vspace{0.1cm}
\noindent\textbf{Acknowledgements.}
This work is supported by the SenseTime-NTU Collaboration Project, Singapore MOE AcRF Tier 1 (2018-T1-002-056), NTU SUG, NTU NAP, the Max Planck-NTU Joint Lab for Artificial Senses and Data Science and Artificial Intelligence Research Lab. We thank Yue Zhao for his participation in discussing the idea.

{\small
\bibliographystyle{ieee_fullname}
\bibliography{bib}
}

\end{document}